\documentclass[runningheads]{llncs2}
\usepackage{graphicx}
\usepackage{amsmath} 
\usepackage{amssymb}  
\usepackage{hyperref}
\usepackage[table]{xcolor}
\usepackage{float}
\usepackage{setspace}
\usepackage{fancyhdr}
\usepackage{multirow}
\usepackage{caption}
\usepackage{subcaption}
\usepackage{dirtytalk}
\usepackage{tabularx}
\usepackage{listings}

\DeclareMathOperator{\EX}{\mathbb{E}}
\begin{document}

\title{MicroRacer: a didactic environment for Deep Reinforcement Learning
}


\author{Andrea Asperti\and Marco Del Brutto}

\institute{
University  of Bologna\\
Department of Informatics: Science and Engineering (DISI)\\
}


\maketitle

\begin{abstract}
MicroRacer is a simple, open source environment inspired by car racing especially meant for the didactics of Deep Reinforcement Learning. The complexity of the environment has been explicitly calibrated to allow users to experiment with many different methods, networks and hyperparameters settings without 
requiring sophisticated software or the need of exceedingly long training times. Baseline agents for major learning algorithms such as DDPG, PPO, SAC, TD2 and DSAC
are provided too, along with a preliminary comparison in terms of training time and 
performance.
\end{abstract}

\section{Introduction}
Deep Reinforcement Learning (DRL) is the new frontier of
Reinforcement Learning \cite{SuttonBook,DRLsurvey,handbookRL}, where Deep Neural Networks are used as function approximators to address the scalability issues of 
traditional RL techniques. This allows agents to make decisions from high-dimensional, unstructured state descriptions without 
requiring manual engineering of input data. The downside of this approach is that learning may require very long trainings,
depending on the acquisition of a large number of unbiased observations of the environment; in addition, since observations are 
dynamically collected by agents, this leads to the well known 
exploitation vs. exploration problem. The need of long training
times, combined with the difficulty of monitoring and debugging the 
evolution of agents, and the difficulty to understand and explain the reasons for possible failures of the learning process, makes 
DRL a much harder topic than other traditional Deep Learning tasks.

This is particularly problematic from a didactic point of view. Most existing environments are either too simple
and not particularly stimulating, like most of the legacy problems
of OpenAIGym \cite{OpenAIGym} (cart-pendulum, downhill slope simulator, \dots), or far too complex, requiring hours
of training (even relatively trivial problems such as those in the Atari family \cite{ALE13,Qlearning15} may
take 12-24 hours of training on a standard laptop, or Colab \cite{COLAB}). Even if, at the end of training, you may observe an advantage of 
a given technique over another, it is difficult to grasp the pros and cons 
of the different algorithms, and forecast their behaviour in different 
scenarios. The long training times make tuning or ablation studies very hard and expensive.
In addition, complex environments are often given in the form of a black-box that essentially prevents event-based monitoring of the evolution of the agent
(e.g. observe the action of the agent in response to a given environment 
situation). Finally, sophisticated platforms like OpenAIgym already offer
state-of-the-art implementations of many existing algorithms; understanding 
the code is complex and time-demanding, frequently obscured by several 
modularization layers (good to maintain but not to understand code);  
as a consequence students are not really induced to put their hands on the code and try personal solutions.

For all these reasons, we created a simple environment explicitly meant for the
didactics of DRL. The environment is inspired by car racing, and has a 
stimulating competitive nature. Its complexity has been explicitly calibrated
to allow students to experiment with many different methods, networks and hyperparameters settings without requiring sophisticated software or 
exceedingly long training times. Differently from most 
existing racing simulation frameworks that struggle in capturing realism, like
Torcs, AWS Deep Racer or Learn-to-race (see Section \ref{sec:related} for a comparison) we do not care for this aspect: one of the important points of the
discipline is the distinction between model-free vs. model-based approaches, 
and we are mostly interested in the former class. From this respect, it is important to communicate to students that model-free RL techniques are supposed to allow interaction with any environment, evolving according to unknown, unexpected and possibly unrealistic dynamics to be discovered by acquiring experience. In the case of MicroRacer, the complexity
of the environment can be tuned in several different ways, changing 
the difficulty of tracks, adding obstacles or chicanes, modifying the acceleration or the timestep.
Another important point differentiating MicroRacer from other car-racing environments
is that the track is randomly generated at each episode, and unknown to the agent, preventing any form
of adaptation to a given scenario (so typical of many autonomous driving competitions).
In addition to the environment, we provide simple baseline implementations of several
DRL algorithms, comprising  DDPG \cite{DDPG}, TD3 \cite{TD3}, PPO \cite{PPO}, SAC \cite{SAC} and DSAC \cite{DSAC}. 

The environment was proposed to students of the course of Machine Learning at the University of Bologna during the past academic year as a possible project for 
their examination, and many students accepted the challenge obtaining interesting
results and providing valuable feedback. We plan to organize a championship for the
incoming year.

The code is open source, and it is available at the following github repository: \href{https://github.com/asperti/MicroRacer}{\url{https://github.com/asperti/MicroRacer}}. Collaboration with other universities and research
groups is more than welcome.

\subsection{Structure of the article}
We start with a quick review of related applications (Section \ref{sec:related}), followed by an introduction to the MicroRacer environment (Section \ref{sec:microracer}). The baseline learning models currently integrated into the systems are discussed in Section \ref{sec:learning models}; their comparative training costs and performances are evaluated in Section \ref{sec:baselines}. Concluding remarks and plans for future research and collaborations are given in Section \ref{sec:conclusions}.

\section{Related Software}\label{sec:related}
We arrived to the decision of writing a new application as a consequence of our dissatisfaction, for the didactic of Reinforcement Learning, of all environments we tested. Several thesis developed under the supervision of the first author \cite{Vorabbi,Galletti} have been devoted to study the suitability of these environments for didactic purposes, essentially leading to negative conclusions. Here, we briefly review some of these applications, closer to the spirit of MicroRacer. 
Many more systems exists, such as \cite{Duckietown,autorally,CARLA}, but they have a strong robotic commitment and a sym2real emphasis that is 
distracting from the actual topic of DRL, and quite demanding in terms of computational resources.

\subsection{AWS Deep Racer}\label{sec:aws}
\href{https://aws.amazon.com/it/deepracer/}{AWS Deep 
Racer}\footnote{\url{https://aws.amazon.com/it/deepracer/}} \cite{AWSDeepRacer} is a cloud based 3D racing simulator developed by Amazon. It emulates a fully autonomous 1/18th scale race car; a global racing league is organized each year.
Amazon only provides utilities to train agents remotely, and with 
very limited configurability: essentially, the user is only able to tune the system of rewards, that gives a wrong didactic message: manipulating rewards is a bad and easily biased way of teaching a behaviour. 
Moreover, at the time it was tested, the AWS DeepRacer console only supported the proximal policy optimization (PPO) algorithm \cite{PPO}; the most recent release should also support Soft Actor Critic \cite{SAC}.

Do to this limitations, a huge effort has been done by the aws-community to pull together the different components required for DeepRacer local training (see e.g. \href{https://github.com/aws-deepracer-community/deepracer-core}{\url{https://github.com/aws-deepracer-community/deepracer-core}}. 

The primary components of DeepRacer are four docker containers:
\begin{itemize}
\item Robomaker Container: Responsible for the robotics environment. Based on ROS + Gazebo as well as the AWS provided "Bundle";
\item Sagemaker Container: Responsible for training the neural network;
\item Reinforcement Learning (RL) Coach: Responsible for preparing and starting the Sagemaker environment;
\item Log-Analysis: Providing a containerized Jupyter Notebook for analyzing the logfiles generated.
\end{itemize}
The resulting platform is extremely complex, computationally demanding, difficult to install and to use.
See \cite{Vorabbi} for a deeper discussion of the limitations of this environment for the didactics of Reinforcement Learning. 

\subsection{Torcs}
\href{https://sourceforge.net/projects/torcs/}{TORCS}\footnote{\url{https://sourceforge.net/projects/torcs/}} is a portable, multi platform car racing simulation environment, originally conceived by
E.Espié and C.Guionneau. It can be used as an ordinary car racing game, or a platform for AI research \cite{Torcs,racing_line}. It runs on Linux, FreeBSD, OpenSolaris and Windows. The source code of TORCS is open source, licensed under GPL. 
While supporting a sophisticated and realistic physical model, it provides a sensibly simpler platform than AWS DeepRacer, and it is a definitely better choice. It does not support random generation of tracks, but many tracks, opponents and cars are available. 

A gym-compliant python interface to Torcs was recently implemented in \cite{Galletti}, under the supervision of the first author. 
While this environment can be a valuable testbench for experts of Deep 
Reinforcement Learning, its complexity and especially the difficulty 
of training agents is an insurmountable obstacle for neophytes. 

\subsection{Learn-to-race}
\href{https://learn-to-race.org/}{Learn-to-Race}\footnote{\url{https://learn-to-race.org/}} \cite{LearnToRace,chen2021safe} is a recent Gym-compliant open-source framework based on a high-fidelity racing simulator developed by Arrival, able to capture complex vehicle dynamics and to render 3D photorealistic views.

Learn-to-Race provides customizable, multi-model sensory inputs giving  information about the state of the vehicle (pose, speed, etc.), and 
comprising RGB image views with semantic segmentations.  
A challenge based on 
Learn-to-Race is organized by AICrowd (similarly to AWS): \href{https://www.aicrowd.com/challenges/learn-to-race-autonomous-racing-virtual-challenge}{\url{https://www.aicrowd.com/challenges/learn-to-race-autonomous-racing-virtual-challenge}}.

Learn-to-race is very similar, in its intents and functionalities, to Torcs
(especially to the gym-compliant python interface developed in \cite{Galletti}). It also shares with TORCS most of the defects: learning 
the environment and training an agent requires a commitment far beyond 
the credits associated with a typical course in DRL; it can possibly be a subject for a thesis, but cannot be used as a didactic tool. 
Moreover, the complexity of the
environment and its fancy (but onerous) observations are distracting
students from the actual content of the discipline.

\subsection{CarRacing-v0}
This is a racing environment available in OpenAI gym. The state consists of a 96x96 pixels top-down view of the track. The action is composed of three continuous values: steering, acceleration and braking. Reward is -0.1 every frame and +1000/N for every track tile visited, where N is the total number of tiles in track. Episode finishes when all tiles are visited. The track is randomly generated at each episode. A few additional indicators at the bottom of the window provide additional information about the car: speed, four ABS sensors, steering wheel position, gyroscope. The game is considered solved when an agent consistently get 900 or more points per episode. As observed in \cite{CarRacing}, the problem is quite challenging due to the peculiar notion of state, that requires learning from pixels: this shifts the focus of the problem from the learning task to the elaboration of the observation, adding a pointless and onerous burden. In addition, while it is
a good practice to stick to a gym-compliant interface for the interaction
between the agent and the environment, for the didactic reasons already explained in the introduction, we prefer to avoid a direct and extensive use of OpenAI gym libraries (while we definitely encourage students to use these
libraries as a valuable source of documentation).

\section{MicroRacer}\label{sec:microracer}
MicroRacer generates new random circular tracks at each episode. The Random track is defined by CubicSplines delimiting the inner and outer border; the number of turns and the width of the track are configurable. From this description, we derive a dense matrix of points of dimension 1300x1300 providing information about positions inside the track. This is the actual definition of the track used by the environment.
The basic track can be further complicated by optionally adding obstacles (similar 
to cars stopped along the track) and \say{chicanes}. More details about the 
environment can be found in \cite{DelBrutto}.

\begin{table}[ht]
\includegraphics[width=\textwidth]{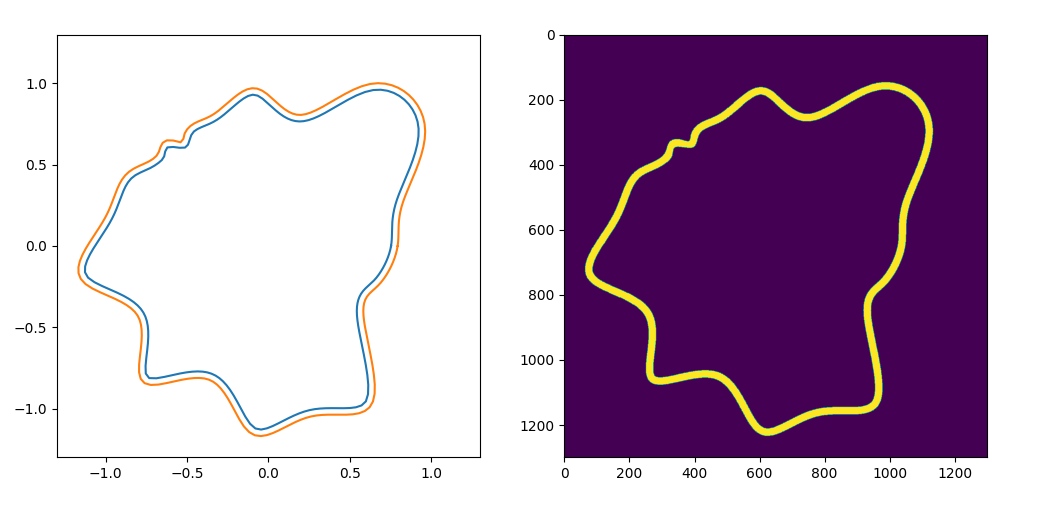}
\caption{(left) Random track generated with splines; (right) derived boolean map. The dynamic of the game is entirely based on the map.
The map is unknown to agents, that merely have agent-centric 
sensor-based observations: speed and lidar-like view.}
\end{table}

\subsection{State and actions}
MicroRacer does not intend to model realistic car dynamics. The model is explicitly meant to be as simple as possible, with the minimal amount of complexity that still makes learning interesting and challenging. The maximum car acceleration, both
linear and angular, are configurable. The angular acceleration is used to constraint
the maximum admissible steering angle in terms of the car speed, forbidding the car
to go too fast.

The state information available to actors is composed by:
\begin{itemize}
\item a lidar-like vision of the track from the car's frontal perspective. This is an array of 19 values, expressing the distance of the car from the track's borders along uniformly spaced angles in the range -30°,+30°.
\item the car scalar velocity.
\end{itemize}
The actor (the car) has no global knowledge of the track, and no information about its absolute or relative position/direction w.r.t. the track\footnote{Our actors exploit a simplified {\em observation} of the state discussed in Section \ref{sec:baselines}}.

The actor is supposed to answer with two actions, both in the range [-1,1]:
\begin{itemize}
\item acceleration/deceleration
\item turning angle.
\end{itemize}
Maximum values for acceleration and turning angles can be configured.
In addition, a simple law depending on a tolerated angular acceleration
(configurable) limit the turning angle at high speeds. This is not meant 
to achieve a realistic behaviour, but merely to force agents to learn
to accelerate and decelerate according to the configuration.

The lidar signal is computed by a simple iterative function written in cython \cite{cython} for the sake of efficiency.

\subsection{Rewards}
Differently from  other software applications for autonomous driving,
shaping rewards from a wide range of data relative to the distance of the car from
borders, deviation from the midline, and so on, \cite{car_rewards,planningcars,AWSDeepRacer}
MicroRacer induces the use of a simple, almost intrinsic \cite{intrinsic04}, rewarding mechanism. Since the objective is to run as fast as possible, it is natural to use speed as the only reward. 
The cumulative reward is thus the integral of speed, namely the expected (discounted) total distance covered by the car. A negative reward is given in case of termination with failure (too slow, or out of borders).
Users are free to 
shape different rewarding mechanisms, but the limited state information
is explicitly meant to discourage this pursuit. It is important for
students to realize that ad-hoc rewards may easily introduce biases in the learning process, inducing agents to behave according to possibly sub-optimal strategies. 

\subsection{Environment interface}
To use the environment, it is necessary to instantiate the \lstinline{Racer} class in \lstinline{tracks.py}.
On initialization, it is possible to turn off obstacles, chicanes, turn and low-speed constraints. The Racer class has two main methods, implementing a OpenAI compliant interface with the environment:
\begin{description}
    \item   \lstinline{reset() -> state}\\
    this method generates a new track and resets the racer position at the starting point. It returns the initial state.
    \item   \lstinline{step(action) -> state, reward, done}\\
    this method takes an action composed by \lstinline{[acceleration, turn]} and lets the racer perform a step in the environment according to the action.
    It returns the new \lstinline{state}, the \lstinline{reward} for the action taken and a boolean \lstinline{done} that is true if the episode has ended.
\end{description}

\subsection{Competitive Race}
In order to graphically visualize a run it is necessary to use the function: 
\begin{lstlisting}
   newrun(actors, obstacles=True, turn_limit=True, 
          chicanes=True, low_speed_termination=True)
\end{lstlisting}
defined in \lstinline{tracks.py}. 
It takes as input a list of \lstinline{actors} (Keras models), simulating a race
between them. At present, the different agents are not supposed to interfere 
with each other: each car is running separately and we merely superpose their trajectories.

\subsection{Dependencies}
The project just requires basic libraries: tensorflow, matplotlib, scipy.interpolate (for Cubic Splines) numpy, and cython. A \verb+requirements+ file is available so you can easily install all the dependencies just using the following command "pip install -r requirements.txt".

\section{Learning models}\label{sec:learning models}
In this section, we list the learning algorithms for which a base code
is currently provided, namely DDPG, TD3, PPO, SAC and DSAC. The code is meant to offer to students a starting point for further development, extending the code and implementing variants. 
All implementations take advantage of {\em target networks} \cite{atari} to stabilize training. 

\subsection{Deep Deterministic Policy Gradient (DDPG)} 
DDPG \cite{Silver14,DDPG} is an off-policy algorithm that 
extends deep Q-learning to continuous
action spaces, jointly learning a Q-function and a policy. It uses off-policy data and the Bellman equation to learn the Q-function, and uses the Q-function to learn the policy. The optimal action-value function $Q^*(s,a)$, and the optimal policy 
$\pi^(s)$ should satisfy the equation
\[Q^*(s,a) = \EX_{s\sim P}r(s,a)+\gamma Q(s',\pi^*(s')\]
that allows direct training of the Q-function from transitions $(s,a,s',r,T)$, similarly to DQN \cite{atari}; in turn, 
the optimal policy is trained by maximazing, over all possible states, 
the expected reward
        \[Q(s,\pi^*(s)\]

\subsection{Twin Delayed DDPG (TD3)} This is a variant of DDPG meant to overcome some shortcomings of this algorithm mostly related to a possible
over-estimation of the Q-function \cite{TD3}. Specifically, TD3 exploits the following tricks:
\begin{enumerate}
\item Clipped Double-Q Learning. Similarly to double Q-learning, 
two "twin" Q-functions are learned in parallel, and the smaller of the two Q-values is used in the r.h.s. of the Bellman equation for computing 
gradients;
\item “Delayed” Policy Updates. The policy (and its target network) is
updated less frequently than the Q-function;
\item Target Policy Smoothing. Noise is added to the target action
inside the Belmman equation, essentially smoothing out Q with respect to changes in action.
\end{enumerate}

\subsection{Proximal Policy Optimization (PPO)} 
A typical problem of policy-gradient techniques is that they 
are very sensitive to training settings: since long trajectories
are into account, modifications to the policy are amplified, 
possibly leading to very different behaviours and numerical instabilities.
Proximal Policy Optimization (PPO) \cite{PPO} simply relies on ad-hoc clipping in the objective function 
to ensure that the deviation from the previous policy is relatively 
small.

\subsection{Soft Actor-Critic (SAC)} Basically, this is a variant
of DDPG and TD3, incorporating ideas of Entropy-regularized Reinforcement
Learning \cite{SAC}. The policy is trained to maximize a trade-off between expected return and entropy, a measure of randomness of the policy. Entropy 
is related to the exploration-exploitation trade-off: increasing 
entropy results in more exploration, that may prevent the policy from prematurely converging to a bad local optimum; in addition, it
add a noise component to the policy producing an effect similar
to Target Policy Smoothing of T3D. It also exploits the clipped 
double-Q trick, to prevent fast deviations. 

\subsection{DSAC}
Distributional Soft Actor-Critic (DSAC) is an off-policy actor-critic algorithm developed by Jingliang et al\cite{DSAC} that is essentially
a variant of SAC where the clipped double-Q learning is substituted by a distributional action-value function \cite{DistributionalRLbook}.
The idea is that learning a distribution, instead of a single value, can help to mitigate Q-function overestimation.
Furthermore, DSAC uses a single network for the action-value estimation, 
with a gain in efficiency.

\section{Baselines benchmarks}\label{sec:baselines}
In this section we compare our baselines implementations
in the case of an environment with a time step of 
$0.04$ ms, and 
comprising obstacles, 
chicanes, low speed termination and turn limitations.

The different learning models are those mentioned in section \ref{sec:learning models}. In the case of  DDPG we 
also considered a variant, called DDPG2
making use of parameter space noise \cite{paramnoiseDDPG}
for the actor's weights.
This noise is meant to improve exploration and it can be used as a surrogate for action noise. 

All models work with a simplified {\em observation} of the environment state, where the full lidar signal is replaced
by 4 values: the angle (relative to the car) of the lidar 
max distance, the value of this distance and the values of 
the distances for the two adjacent positions. In mathematical 
terms, if $\ell$ is the vector of lidar signals, $m = argmax(\ell)$
and $\alpha_m = angle(m)$ is the corresponding direction, the ovservation is composed by
\[\alpha_m,\ell(m-1),\ell(m),\ell(m-1)\]


The DDPG actor's neural network makes use of two towers. 
One of them calculates the direction, while the other calculates the acceleration. 
Each of them is composed of two hidden layers of 32 units, with relu activation.
The output layer uses a tanh activation for each action.
At the same time, the critic network uses two layers, one of 16 units and one of 32, for the state input and one layer of 32 units for the action input.
The outputs of these layers are then concatenated and go through another two hidden layers composed of 64 units.
All of them make use of relu activation.

In DDPG2, the actor has two hidden layers with 64 units and relu activation and one output with tanh activation .
Meanwhile, the critic is the same as in DDPG.

In TD3, the actor is the same as DDPG2.
The critic has two hidden layer with 64 units and relu activation.

In SAC, the actor has two hidden layer with 64 units each and relu activation and output a $\mu$ and a $\sigma$ of a normal distribution for each action.
The critic is equal to TD3.

In DSAC, the actor is the same as SAC.
The critic has the same structure as the actor.

In PPO both the actor and the critic have two hidden layers of 64 units with tanh activation, but the actor has also tanh activation on the output layer.

All learning methods have been trained with a discount factor $\gamma=0.99$, using Adam as optimizer. 
All methods except PPO share the following hyperparameters:
\begin{itemize}
\item[-] Actor and Critic Learning Rate 0.001
\item[-] Buffer Size 50000
\item[-] Batch Size 64
\item[-] Target Update Rate $\tau$ 0.005
\end{itemize}
Additional methods-specific hyperparameters are listed in Table \ref{table:hyperparams}.
\begin{table}[H]
\centering
\begin{tabularx}{\textwidth}{ll}
\begin{tabular}[t]{@{}ll@{}}
\hspace{0.4cm}\textbf{Hyperparameter}    & \textbf{Value} \\ \hline\\
\textit{TD3, DDPG}               &                \\
\hspace{0.4cm} Exploration Noise          &    $\mathcal{N}(0, 0.1)$   \\
\textit{TD3}               &                \\
\hspace{0.4cm} Target Update Frequency    & 2              \\
\hspace{0.4cm} Target Noise Clip          & 0.5            \\
\textit{SAC, DSAC}         &                \\
\hspace{0.4cm} Target Entropy             & -$A$           \\
\textit{DSAC}              &                \\
\hspace{0.4cm} Target Update Frequency    & 2              \\
\hspace{0.4cm} Minimum critic sigma       & 1              \\
\hspace{0.4cm} Critic difference boundary & 10             \\
\end{tabular} &
\begin{tabular}[t]{@{}ll@{}}
\hspace{0.4cm}\textbf{Hyperparameter}    & \textbf{Value} \\ \hline\\
\textit{DDPG2}             &                \\
\hspace{0.4cm} Parameter Noise Std Dev    & 0.2            \\
\textit{PPO}               &                \\
\hspace{0.4cm} Actor/Critic Learning Rate        & 0.0003          \\
\hspace{0.4cm} Mini-batch Size                 & 64             \\
\hspace{0.4cm} Epochs                 & 10             \\
\hspace{0.4cm} GAE lambda                 & 0.95             \\
\hspace{0.4cm} Policy clip                 & 0.25            \\
\hspace{0.4cm} Target entropy                 & 0.01             \\
\hspace{0.4cm} Target KL                           &   0.01 
\end{tabular}
\end{tabularx}
\caption{Hyperparameters used in the various methods.}
\label{table:hyperparams}
\end{table}

\subsection{Results}
Training times have been computed as an avarage of ten different trainings, each one conssisting of 50000 training steps. In the case of PPO, that unlike all the other methods, starts collecting a complete trajectory before executing a training step
on it, we trained the agent for a fixed number of episodes (600).

The training times collected are relative to the execution on two different machines: a laptop equipped with an NVIDIA GeForce GTX 1060 GPU, Intel Core i7-8750H CPU and 16GB 2400MHz RAM, 
and a wokstation equipped with an Asus GeForceDUALGTX1060-O6G GPU and a Intel Core i7-7700K CPU and 64GB 2400 MHz RAM..

As can be observed in Table \ref{table:traininttime}, the methods that train an higher number of Neural Networks require higher training times.
\begin{table}[H]
\centering
\begin{tabular}{|l|l|l|l|l|l|l|}
\hline
\textbf{\textbf{Machine}} & DDPG & DDPG2 & TD3   & SAC  & DSAC   & PPO \\ \hline
\textbf{M1}               & 30m  & 44m   & 38m   &19m   & 24m  &  27m   \\ \hline
\textbf{M2}               & 14m  & 24m   & 23m   &11m   & 12m  &  20m   \\ \hline
\end{tabular}%
\caption{Average training time (5 runs) required to perform 50000 training iterations (600 episodes for PPO) for each different method. Times are relative to two different machines: M1 is a laptop equipped with an NVIDIA GeForce GTX 1060 GPU, Intel Core i7-8750H CPU and 16GB 2400MHz RAM, M2 is a workstation equipped with an Asus GeForceDUALGTX1060-O6G GPU and a Intel Core i7-7700K CPU and 64GB 2400 MHz RAM.}
\label{table:traininttime}
\end{table}

\begin{figure}[H]
    \centering
    \includegraphics[width=.9\textwidth]{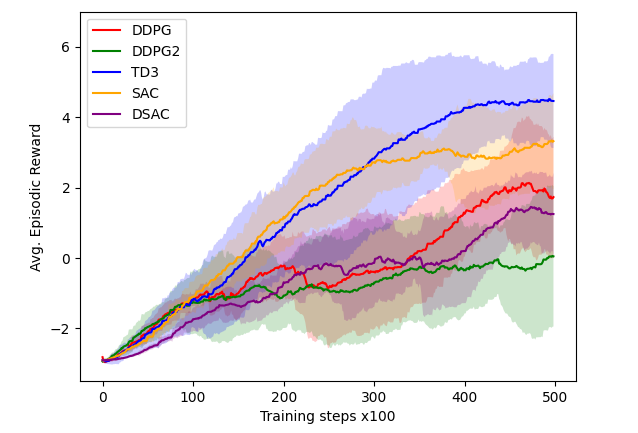}
    \caption{Training curves of all methods except PPO. The solid lines correspond to the mean and the shaded regions correspond to 95\% confidence interval over 10 trainings.}
    \label{fig:trainingcurves}
\end{figure}

As can be seen in Figure \ref{fig:trainingcurves}, the training process has 
large fluctuations, also due to frequent occurrences of catastrophic forgetting
(more on it below).
TD3 and SAC are the most stable methods, usually requiring less observations and training steps to improve. 
The other methods learns at a slower pace and seem to be more prone to catastrophic forgetting. However, they are occasionally able to produce reasonably performant agents.

After each training, 100 evaluation episodes has been run to collect the real performance of each trained agents.
The average of these results over 10 different trainings and the best results obtained for each method can be seen in Table \ref{table:evalaverage}.
As it can be noticed, a higher number of completed episodes usually corresponds to slower speeds.
This may indicate difficulties in the process of learning the right acceleration action.
Similarly to the training curves, TD3 and SAC seem to have the best performances even in evaluation, as expected.

\begin{table}[H]
\centering
\resizebox{\textwidth}{!}{%
\begin{tabular}{|l|l|l|l|l|l|l|}
\hline
\textbf{Method}                     & DDPG & DDPG2 & TD3  & SAC & DSAC & PPO \\ \hline
\textbf{Average completed episodes} & 38   & 18    & 54   & 69  & 37   &  37   \\ \hline
\textbf{Average episodic reward}    & 2.48 & 0.80  & 3.52 & 4.61& 2.84 &   2.05  \\ \hline
\textbf{Average speed}              & 0.34 & 0.30  & 0.26 & 0.29& 0.34 &   0.23  \\ \hline
\textbf{Max completed episodes}     & 90   & 39    & 80   & 79  & 75   &   62  \\ \hline
\end{tabular}%
}
\caption{Average and maximum of 100 evaluation episodes executed after each training over 5 trainings of 50000 iterations (600 episodes for PPO). }
\label{table:evalaverage}
\end{table}

\begin{figure}[H]
    \centering
    \includegraphics[width=0.49\textwidth]{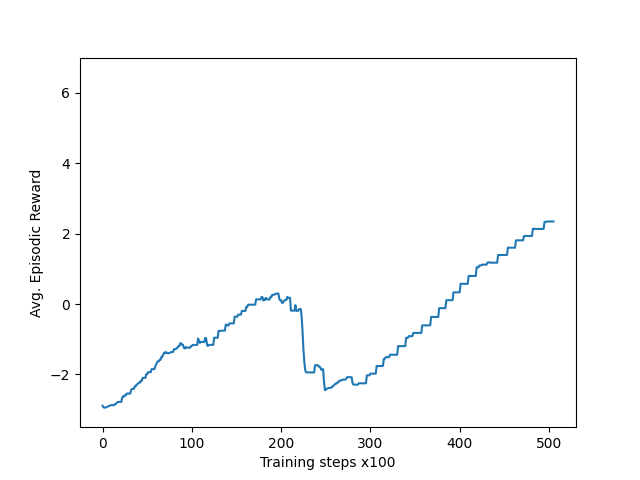}
    \includegraphics[width=0.49\textwidth]{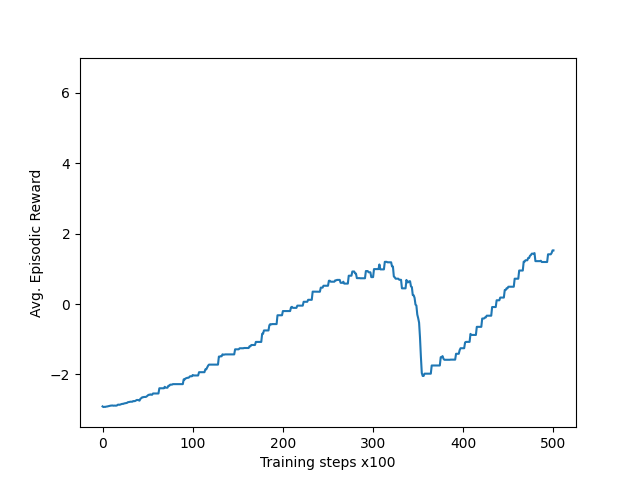}
    \caption{Examples of catastrophic forgetting during training of 
    DDPG (left) and DSAC (right).}
    \label{fig:catastrophe}
\end{figure}

In Figure \ref{fig:catastrophe} we show a few examples of catastrophic forgetting, that is the tendency of a learning model to completely and abruptly forget previously learned information during its training. The phenomenon is still largely misunderstood, so having a relatively simple and highly configurable environment where we can frequently observe its occurrence seems to provide a very interesting and promising framework for future investigations.

\vspace{-1cm}
\section{Conclusions}\label{sec:conclusions}
In this article, we introduced the MicroRacer environment, offering a simple educational platform for the didactic of Reinforcement Learning. Similarly
to our previous environment based on the old and prestigious Rogue game \cite{RogueLOD,RogueIEEE}, we 
try to spare the useless burden of relying on two-dimensional state observations requiring expensive image-preprocessing, using instead more direct and synthetic state information. Moreover, differently from Rogue, 
that was based on a discrete action-space, MicroRacer is meant to investigate RL-algorithms with continuous actions. 

On the contrary of most existing car-racing systems, MicroRacer does not make any attempt to implement realistic dynamics: autonomous driving is just a simple pretext to create a pleasant and competitive setting. This drastic simplification
allows us to obtain an environment that, although far from trivial, still has acceptable training times (between 10 and 60 minutes depending on the learning methods and the underlying machine).

The environment was already experimented by students of the course of Machine Learning at the University of Bologna during the past academic 
year, that provided valuable feedback. In view of the welcome reception,
we plan to organize a championship for the incoming year.
The code is open source, and it is available at the following github repository: https://github.com/asperti/MicroRacer. We look forward for possible collaborations with other Universities and research institutions.

\bibliographystyle{plain}
\bibliography{machine.bib,DRL.bib}

\end{document}